\pdfoutput=1
\documentclass[manuscript,screen]{acmart}

\usepackage{booktabs}
\usepackage{array}
\usepackage{multirow}
\usepackage{graphicx}
\usepackage{url}
\usepackage{xspace}
\usepackage{booktabs}
\usepackage{placeins}
\usepackage{tabularx}
\usepackage{ragged2e}

\AtBeginDocument{%
  
}

\setcopyright{none}
\copyrightyear{2026}
\acmYear{2026}
\acmDOI{}
\acmConference[Manuscript Submitted to ACM]{Manuscript submitted to ACM}{April 2026}{ }
\acmISBN{}

\begin{document}
\title{Assessing the Performance of AI Models in Misinformation Detection Using Predefined Social Noise Constructs}

\author{Manar Alsaid}
\affiliation{%
\institution{East Texas A\&M University}
\department{Department of Computer Science and Information Systems}
\city{Commerce}
\state{Texas}
\country{USA}
}
\email{manar.alsaid@etamu.edu}

\ccsdesc[500]{Information systems~Social networks}
\ccsdesc[500]{Computing methodologies~Natural language processing}
\ccsdesc[500]{Computing methodologies~Machine learning}

\begin{abstract}
Social media platforms generate vast volumes of user generated content that often obscure meaningful discourse, a phenomenon conceptualized as social noise. While existing misinformation detection methods primarily focus on factual verification, they largely overlook the role of social noise in amplifying misinformation. This study assesses the performance of 6 text-based models, in addition to GNN, in mitigating the effects of misinformation using predefined social noise constructs. The dataset consists of 500,000 tweets collected from the X platform using the \#Ukraine hashtag during the Russia-Ukraine conflict from March 2024 to March 2025. Advanced machine learning models achieve strong performance, with accuracies ranging from 79.94\% to 97.17\%. RoBERTa demonstrates the highest performance (accuracy = 97.17\%, Cohen's $\kappa = 0.9660$, MCC = 0.9660, AUC = 0.9977). GNNs on the other hand which exploit the relational structure between tweets have shown accuracy of with 64.52\% and $\mathrm{AUC} = 0.899$ achieving well above random chance classification graph topology. while all models achieve $\kappa > 0.75$, indicating substantial overall agreement. Explainable AI analysis using LIME reveals that 89.6\% of predictive weight is attributed to contextual features rather than isolated keywords. The findings highlight that integrating social noise into AI pipelines enhances misinformation detection and supports more effective mitigation strategies in large-scale social media environments. 

\end{abstract}

\keywords{Social noise, machine learning, deep learning, exploratory data analysis, explainable AI, LIME, natural language processing}

\maketitle

\section{Introduction}

Social media platforms have become primary channels for public discourse, news dissemination, and opinion formation. However, the sheer volume of user-generated content and the dynamics of platform engagement make these informational environments highly susceptible to noise and informational distortion. A significant portion of this dilution arises from what is conceptualized as \emph{social noise}: content that obscures meaningful signals and complicates the interpretation of online discourse \citep{madali2024socialnoise}.

While misinformation and disinformation are concerned with misleading and false content, social noise represents a broader category of communication that is not necessarily correct or false but rather out of context, irrelevant, performative, or misaligned with the original intended message. Understanding the motives driving the generation of such content is critical in combating the spread of misinformation and disinformation. Social noise manifests in various forms, including performative identity expressions, ritualized engagement behaviors, image curation practices, affiliation signaling, and formulaic displays of solidarity. Collectively, these patterns can dominate social media feeds, amplifying visibility without necessarily improving comprehension of the events or issues being discussed \citep{madali2024socialnoise,zhao2026activelearning}.

Efforts in misinformation detection have primarily focused on assessing the factual accuracy of claims by comparing them against verified external knowledge sources~\cite{10.1145/3812649,10.1145/3363818}. In contrast, \textit{social noise} refers to content that is not necessarily false, but contributes little substantive informational value within a given discourse. For example, a tweet expressing formulaic support, such as \texttt{\#StandWithUkraine}, without additional context, does not constitute misinformation. However, when produced at scale, such content can dilute the visibility of credible reporting, nuanced analysis, and firsthand accounts. This distinction is both theoretically and operationally significant because systems optimized solely for factual verification are insufficient to capture the broader degradation of informational quality caused by social noise.

Prior studies on social noise have shown that user behavior on social media is shaped by social, psychological, and contextual factors that influence how information is produced, shared, and amplified. These studies identify multiple constructs as shown in Appendix A, including image curation, affiliation signaling, and relationship management, that motivate participation in social noise and may indirectly contribute to the spread of misinformation \citep{madali2024socialnoise, alsaid2022social, zimmerman2022socialdoc}. Emerging work on explainable misinformation detection also emphasizes the importance of distinguishing between surface-level textual cues and semantically meaningful patterns in social media data \citep{ngueajio2025xai}.

Machine learning and deep learning models could be used to help curb misinformation by classifying large amounts of data and identifying motives behind the generation of irrelevant and out-of-context content, such as social noise\citep{chen2026jobpocalypse}. Text-based classification using deep learning has advanced rapidly with the introduction of pre-trained language models. BERT and its variants, including RoBERTa, DeBERTa, and ALBERT, have achieved strong results across NLP benchmarks by leveraging bidirectional contextual representations learned from large-scale pretraining \citep{devlin2019bert,liu2019roberta}. RoBERTa, which is employed in this study, improves upon BERT by introducing dynamic masking, larger batch sizes, and removing the next-sentence prediction objective \citep{liu2019roberta}. The GNN models add a network-science perspective that validates the construct taxonomy at a structural level. They introduce propagation dynamics as an analytical dimension, and demonstrate the application of cutting-edge graph-based deep learning (GIN, GAT) to the social noise classification problem. This broadening of the analytical toolkit from six text based classifiers to ten total models (six text-based + four graph-based) across two fundamentally different modeling paradigms and strengthens the analysis's contribution as a comprehensive, multi-method analysis of social noise on social media.

Explainable AI (XAI) techniques, such as LIME and SHAP, are increasingly used to enhance transparency and interpretability in model predictions. These methods enable researchers to identify the linguistic and contextual features that drive classification decisions, thereby improving trust and supporting theoretical validation. However, their effectiveness remains contingent on model design and feature dependencies, making careful interpretation necessary.

\section{Related Work}

Existing research on misinformation and disinformation detection has predominantly focused on factual versus non-factual information, with little attention to the gray areas where arguments can be debated \citep{glockner2026evidence} \citep{walter2026detecting},\citep{alsaid2024combating}, \citep {ahvanooey2022misinformation}. In contrast, the concept of social noise captures the informational utility of content within a discourse ecosystem. Social noise reflects communicative behaviors driven by social and psychological motivations, such as affiliation signaling, image curation, and normative expression, rather than truthfulness alone. For instance, a tweet expressing support through a hashtag may be factually accurate but contribute limited informational value, thereby increasing noise and reducing the visibility of substantive content.

This distinction is critical: misinformation detection addresses veracity, whereas social-noise classification captures communicative intent and discourse impact. Emerging evidence suggests that high levels of social noise can obscure credible information, amplify polarization, and create conditions that facilitate the spread of misinformation \citep{ngueajio2025xai},\citep{madali2024socialnoise}. 

The computational study of misinformation has matured rapidly over the past
decade. Vosoughi et al.\citep{vosoughi2018spread} using a corpus
of roughly 126{,}000 have demonstrated that falsehoods information and rumors diffuse
``farther, faster, deeper, and more broadly'' than the truth, motivating a large
body of automated detection research. Shu et al.\citep{shu2017fakenews} using different approach formalized fake-news
detection as a data-mining task organized around news content and social
context which rather close to the use of the social noise concept. At the same time Zhou and Zafarani\citep{zhou2020survey} synthesized fundamental
theories and detection methods across knowledge, style, propagation-, and
source-based perspectives. Further Lazer et al.\citep{lazer2018science} and Pennycook and
Rand\citep{pennycook2021psychology} situated the misinformation phenomenon within its
social and cognitive context, emphasizing that veracity judgments are mediated by
attention, prior belief, and partisan identity rather than factual content
alone. Their approach is similar to the general definition of social noise adopted in this study. 

Early credibility work of misinformation by Castillo et al.\citep{castillo2011credibility}
showed that surface and propagation features on Twitter are predictive of
perceived information credibility. Most detection systems target factual veracity by
comparing claims against external knowledge or by modeling linguistic deception
cues\citep{shu2017fakenews,zhou2020survey}, an emphasis consistent with prior
work in this area by \citep{deng2026crossmodal},\citep{hochreiter1997lstm},\citep{monti2019geometric}. The problem with these approaches is that they treat content as either true or false and largely overlooking the impact of social noise impact on communication behavior and its role in diluting the content of the intended message. 

On the communication behavior aspects, research that examines how communicative behavior, rather than
veracity alone is carried by Zimmerman \citep{zimmerman2022socialdoc}
characterizes \emph{social noise} as the influence of observers on social media
information behavior, and Madali et al.\citep{madali2024socialnoise} and Alsaid and
Madali \citep{alsaid2022social}. These studies operationalized the concept of social noise and conceptualized a framework that consisted of six construct including image curation, affiliation signaling, and
relationship management. Alsaid et al.\citep{alsaid2024combating} extend this
work by introducing social entropy as a measure of discourse uncertainty. This
body of work argues that performative, ritualized, and identity-signaling content
can dominate a feed without being false, thereby reducing the visibility of
substantive reporting. Despite its conceptual development, the systematic \emph{computational} classification of social noise remains largely unexplored, which is the gap this study addresses.

Researchers have explored the use of deep learning to detect offensive, abusive, or misleading content on social media platforms. Bidirectional Encoder Representations from Transformers (BERT) has become a foundational model in natural language processing due to its ability to capture bidirectional context through the Transformer architecture \citep{devlin2019bert}, \citep{xiao2023challenges}. BERT is a pre-trained model designed to understand contextual relationships between words in a sentence and serves as a foundation for numerous NLP applications. For example, \citep{kim2022garlic} used BERT to classify misinformation about garlic and COVID-19 on Twitter. Their results showed that BERTweet-large achieved strong performance, with a maximum accuracy of 0.911 and an F1 score of 0.894.

\citep{uddin2025bertfake} tested several models, including XGBoost, CNN-RNN, BERT, and RoBERTa with graph neural networks, to detect fake news using FakeNewsNet, PolitiFact, and Kaggle fake-news datasets. Their results indicated that RoBERTa with graph neural networks achieved the highest accuracy, outperforming BERT, CNN-RNN, and XGBoost. Recent advances in NLP have also seen the emergence of large language models as powerful tools for text classification. However, the computational cost of LLMs remains prohibitive for large-scale classification tasks involving hundreds of thousands of samples. Moreover, comparative studies have shown that traditional machine learning and deep learning models can remain competitive in domain-specific classification tasks.

\citet{zhang2018graph} proposed an end-to-end deep learning architecture for graph classification. Related geometric deep learning approaches have used content features, user profiles, activity patterns, social graphs, and temporal propagation dynamics to assess credibility and detect fake news on social media \citep{monti2019geometric, 10.1145/3363818}. Hybrid classification approaches have also been studied. For example, \citet{vinnela2025hlmbxr} combined BERT, XGBoost, Random Forest, and logistic regression into a hybrid learning model for spam detection, achieving higher accuracy than standalone classifiers.
The combination of BERT and LSTM has also been examined. BERT extracts rich semantic embeddings through transformer-based self-attention, while LSTM networks model sequential dependencies and long-range contextual information \citep{hochreiter1997lstm}. Prior studies suggest that BERT-LSTM hybrids can improve performance in selected classification tasks by combining broad contextual representation with sequential modeling \citep{wang2025bertlstm},\citep{si2023sentiment}, \citep {zhou2024multiscale}.

The pretrained-transformer paradigm, introduced by the self-attention
architecture of Vaswani et al.\cite{vaswani2017attention} and instantiated for
language understanding by BERT \cite{devlin2019bert}, has reshaped text
classification. RoBERTa \cite{liu2019roberta} strengthens BERT approach through dynamic
masking, larger batches, and the removal of the next-sentence-prediction
objective. These models have been applied to misinformation detection by fine-tune BERT variants to classify
COVID-19 misinformation\citep{kim2022garlic}. additonally, Uddin et al.\cite{uddin2025bertfake}Combined RoBERTa with graph neural networks for fake-news detection. Recurrent
and convolutional architectures remain competitive baselines in which long short-term
Memory networks \cite{hochreiter1997lstm} model sequential dependencies, and hybrid BERT-LSTM designs have improved performance on selected classification tasks. The custom architectures evaluated here build on established components, including pretrained GloVe embeddings~\cite{pennington2014glove}, residual
connections~\cite{he2016resnet}, and squeeze-and-excitation channel
attention~\cite{hu2018senet}, alongside the gradient-boosted trees of
XGBoost~\cite{chen2016xgboost} and the bagged ensembles of Breiman's random
forests~\cite{breiman2001randomforests}. Comparative studies consistently
report that, for domain-specific tasks with strong lexical signals, traditional
Machine-learning models can approach transformer-level performance, a finding that the study results corroborate.

Due to the fact that misinformation and social noise propagate relationally, graph-based models have
become have become an attractive option for the detection of social noise and misinformation. The modern GNN toolkit comprises spectral
convolution (GCN)~\cite{kipf2017gcn}, attention-based aggregation
(GAT)~\cite{velickovic2018gat}, inductive neighborhood sampling
(GraphSAGE)~\cite{hamilton2017graphsage}, and the maximally expressive
Weisfeiler Leman-aligned GIN~\cite{xu2019gin}, Wu et
al.~\cite{wu2021gnnsurvey}  These architectures
underpin the four models we evaluate on the propagation graph. In the misinformation domain, Monti et al.~\cite{monti2019geometric} pioneered
geometric deep learning for fake-news detection where Bian et
al.~\cite{bian2020bigcn} introduced bi-directional GCNs as a
top-down propagation and bottom-up dispersion model of rumors. Social-context models
such as FANG~\cite{nguyen2020fang} learn inductive representations of the
user news graph, while UPFD~\cite{dou2021upfd} fuses endogenous user preference
with an exogenous propagation structure. Graph classification frameworks more
broadly, such as the end-to-end architecture of Zhang et
al.~\cite{zhang2018graph}, inform our treatment of tweet-level
graphs. 

As detection systems are deployed in moderation pipelines, interpretability has
become a requirement rather than an add-on~\cite{arrieta2020xai}. Model-agnostic
methods such as LIME~\cite{ribeiro2016lime} and SHAP~\cite{lundberg2017shap},
together with gradient-based attribution methods such as integrated
gradients~\cite{sundararajan2017ig}, allow researchers to understand and attribute predictions and reasoning of their results 
to specific linguistic and contextual features. Explainability has been
integrated directly into misinformation architectures. dEFEND~\cite{shu2019defend}
identifies check-worthy sentences and user comments via co-attention,
and GCAN~\cite{lu2020gcan} produces graph-aware co-attention explanations over
source tweets and retweeters. In a recent survey by Ngueajio et
al.\cite{ngueajio2025xai}, emphasize the importance of
distinguishing surface-level lexical cues from semantically meaningful patterns
and caution that explanation fidelity is contingent on model design and the ability to understand what happened algorithmeticlly \cite{arrieta2020xai}. The study used LIME for the text models and
gradient-based attribution for the GNNs in order to test whether classification relies on labeling keywords or on broader contextual structure.

This study evaluates a range of machine learning and deep learning architectures, including RoBERTa, BiLSTM with attention, convolutional models, and ensemble methods, for social-noise classification. By operationalizing theoretically grounded constructs and conducting comparative performance analysis, this work extends existing research beyond veracity detection toward a broader understanding of information quality in social media environments. Given that prior work has focused on (i) advanced veracity-oriented misinformation detection~\cite{vosoughi2018spread,shu2017fakenews,zhou2020survey}, (ii)
conceptualized social noise without large-scale computational validation~\citep{devlin2019bert},\citep{wang2025bertlstm},\citep{lundberg2017shap}, (iii) established a strong transformer and graph architectures~\cite{vaswani2017attention},\citep{xu2019gin}, and (iv) developed mature
explainability tooling~\cite{ribeiro2016lime},\citep{shu2019defend}, This work extent prior studies to assess using  systematic, theory-grounded, and explainable evaluation of whether
contemporary AI models have the ability to mitigate the impact of misinformation using predefined social noise constructs. The present study addresses this existing gap in the literature through a $6\times6$ comparison of transformer, deep-learning, and ensemble models, a complementary four-architecture GNN propagation analysis, and a keyword-ablation explainability protocol that together test the robustness and the semantic grounding of the construct distinctions. Also our use of GNNs differs in away that rather than maximizing detection accuracy, we use graph topology as an independent information source to triangulate whether the six social-noise constructs are structurally distinct.

\section{Methodology}

In this study, we used a 6 by 6 evaluation method in which 6 text-based advanced machine learning and deep learning algorithms and models spanning three methodological families. 1 pre-trained transformer (RoBERTa), 2 custom deep learning architectures (BiLSTM with multi-head self-attention and 1 multi-scale residual CNN with squeeze-and-excitation attention), and 3 traditional machine learning ensemble methods (XGBoost, Random Forest, and AdaBoost) evaluated against 6 social noise constructs identified in previous research \citep{madali2024socialnoise}. The 6 models and 6 constructs are shown in Table 1 below.  This selection enables systematic comparison across model complexity, representational capacity, and training paradigms. Social noise, however, is fundamentally relational. Narratives propagate through clusters of semantically similar posts, and a tweet's structural position within that propagation network carries information about its construct label that is orthogonal to the textual features alone. To capture this relational dimension and test whether the six constructs are structurally distinct, we applied four Graph Neural Network (GNN) architectures to a tweet-level propagation graph built from the dataset. 

\begin{table}[h]
\centering
\caption{Six Models Evaluated Across Six Social Noise Constructs}
\label{tab:social_noise_models}
\renewcommand{\arraystretch}{1.2}
\setlength{\tabcolsep}{5pt}
\begin{tabular}{lcccccc}
\toprule
\textbf{Model} &
\textbf{Image} &
\textbf{Relationship} &
\textbf{Conflict} &
\textbf{Cultural} &
\textbf{Affiliation} &
\textbf{Norms} \\
 &
\textbf{Curation} &
\textbf{Management} &
\textbf{Engagement} &
\textbf{Agency} &
\textbf{\& Politics} &
\textbf{\& Beliefs} \\
\midrule
RoBERTa (Pre-trained Transformer) & \checkmark & \checkmark & \checkmark & \checkmark & \checkmark & \checkmark \\

BiLSTM + Multi-head Self-Attention & \checkmark & \checkmark & \checkmark & \checkmark & \checkmark & \checkmark \\

Multi-scale Residual CNN + SE Attention & \checkmark & \checkmark & \checkmark & \checkmark & \checkmark & \checkmark \\

XGBoost & \checkmark & \checkmark & \checkmark & \checkmark & \checkmark & \checkmark \\

Random Forest & \checkmark & \checkmark & \checkmark & \checkmark & \checkmark & \checkmark \\

AdaBoost & \checkmark & \checkmark & \checkmark & \checkmark & \checkmark & \checkmark \\
\bottomrule
\end{tabular}
\end{table}
\FloatBarrier

\subsection{Dataset and Data Preparations}

The dataset comprises 500,000 tweets collected using the \#Ukraine hashtag during the Russia-Ukraine conflict, spanning the period from March 2024 to March 2025. The X API was used to collect the dataset from the X platform. The decision to focus on this hashtag was based on the abundance of misinformation and disinformation associated with competing Western and Eastern narratives about the war in Ukraine.

The data-engineering pipeline applied a sequence of text-normalization steps to prepare tweets for model input. Non-textual information, such as images and URLs, was removed. Analyses performed in English included tokenization, lowercasing, and stop-word removal. The large-scale dataset provided sufficient examples for training machine learning and deep learning models. The data were split into training and test partitions using stratified sampling to preserve class balance across the six social-noise constructs.

\subsection{Model Implementations} 
\subsubsection{RoBERTa}
It is a transformer-based language model pre-trained on a large corpus of English text using a masked language modeling objective \citep{liu2019roberta}. The implementation uses the RoBERTa-base architecture with 124.6 million parameters, consisting of 12 transformer layers, 768 hidden dimensions, and 12 attention heads. The model was fine-tuned on the training set for five epochs with a learning rate of $2 \times 10^{-5}$ using the AdamW optimizer with linear warmup over the first 10\% of training steps.

\subsubsection{BiLSTM with multi-head self-attention.}
The proposed architecture integrates bidirectional long short-term memory (BiLSTM) networks with a multi-head self-attention mechanism to jointly model sequential dependencies and global contextual interactions in tweet data. The model contains approximately 8.2 million trainable parameters. Input tokens are represented using pre-trained 200-dimensional GloVe embeddings, which are kept fixed during training to mitigate overfitting.

The embedding layer feeds into two stacked BiLSTM layers, each with 256 hidden units per direction, generating contextualized token representations that capture both preceding and succeeding information within the sequence. On top of these representations, a multi-head self-attention module with eight attention heads is applied, where each head computes scaled dot-product attention over 64-dimensional projections. This design enables the model to learn diverse contextual relationships across tokens from multiple representational subspaces.

\subsubsection{Multi-scale residual CNN with squeeze-and-excitation attention.}
The proposed multi-scale residual convolutional neural network (CNN) architecture leverages parallel convolutional streams to extract hierarchical $n$-gram features, complemented by squeeze-and-excitation (SE) blocks for adaptive channel-wise feature recalibration. The model comprises 9.5 million trainable parameters. To capture linguistic patterns across multiple granularities, the input is processed through three parallel branches with kernel sizes of 3, 5, and 7, respectively targeting trigram, 5-gram, and 7-gram dependencies. Each branch is composed of two convolutional layers with 256 filters, integrated with batch normalization and ReLU activation functions. Residual connections are implemented across each block to facilitate gradient flow and support stable deep feature extraction.

Feature refinement is further achieved through the integration of SE modules following each convolutional stage. These blocks first aggregate spatial information through global average pooling to generate a channel descriptor. This is followed by an excitation phase using a two-layer fully connected bottleneck with a reduction ratio of $r=16$. A final sigmoid activation produces per-channel scaling factors, allowing the network to emphasize salient feature maps while suppressing redundant or non-discriminative information across the multi-scale representations.

\subsubsection{Extreme Gradient Boosting (XGBoost).}
It is an optimized implementation of gradient-boosted decision trees, which was selected for its robust performance in high-dimensional classification tasks. The model was configured with 1{,}000 boosting iterations and a maximum tree depth of eight. To maintain generalization and mitigate overfitting, the learning rate was set to 0.1, using a subsample ratio and a column-wise sampling ratio of 0.8 per tree. Further regularization was enforced through a minimum child weight of 5 and an $L_2$ regularization parameter ($\lambda$) of 1.0.

The feature space comprised TF-IDF vectors derived from unigram, bigram, and trigram tokens, constrained to a maximum vocabulary of 50{,}000 features. The training objective used a multi-class softmax loss function across six output classes. XGBoost's efficacy stems from its sequential ensemble strategy, in which successive learners minimize residual errors from the preceding ensemble. Unlike traditional first-order methods, XGBoost leverages second-order Taylor expansion through gradients and Hessians in its split-finding algorithm, enabling a more precise approximation of the loss surface.

\subsubsection{Adaptive Boosting (AdaBoost).}
The AdaBoost model was implemented with 500 estimators, each a decision tree with a maximum depth of 6. The implementation used the SAMME.R algorithm with a learning rate of 0.5, which leverages weighted probability estimates over discrete class predictions. Following the standard adaptive procedure, the model sequentially reweights training samples, forcing the ensemble to concentrate representational capacity on previously misclassified observations. Despite its shared identity with other boosting-based approaches, AdaBoost achieved an accuracy of 79.94\%, substantially lower than XGBoost.

\subsubsection{Random Forest.}
The Random Forest classifier was configured with an ensemble of 1{,}000 decision trees, each trained through bootstrap aggregation with randomized feature selection at each node split. While maximum tree depth was unconstrained, regularization was enforced by requiring a minimum of two samples per leaf and five samples to initiate a split. Following established heuristics for high-dimensional classification, the number of features considered at each split was limited to the square root of the total feature dimensionality. The model used the same 50{,}000-dimensional TF-IDF vectors as XGBoost, with final predictions determined through majority voting across the ensemble.

Random Forests derive robustness from the diversity of individual learners, achieved through data and feature subsampling. Unlike gradient boosting methods, which construct trees sequentially to minimize residual error, Random Forest trees are generated independently and in parallel. This architectural distinction accounts for the observed performance difference between Random Forest and XGBoost.

\subsection{Explainability Analysis}

Local Interpretable Model-agnostic Explanations (LIME) were used to audit the text
classification models and verify whether predictions were based on contextual semantics or superficial keyword
matching. The top 20 contributing features were extracted and categorized for construct-defining keywords. And for non-keyword contextual features. A higher
concentration of non-keyword features serves as an indicator of genuine contextual learning. This analysis was
performed on 2,000 randomly sampled Tweets for each model. The non-keyword the 
weight percentage was computed by dividing the sum of absolute LIME coefficients for non-keyword features by the
total absolute coefficient sum, providing a direct metric for assessing model reliance on broader linguistic context.

\subsection{Graph Neural Network Propagation Analysis}
The six classifiers described in the preceding sections, RoBERTa, the BiLSTM with multi-head self-attention, the multi-scale residual CNN, XGBoost, Random Forest, and AdaBoost, each treat a tweet as an independent observation. Social noise, however, is fundamentally relational. Narratives propagate through clusters of semantically similar posts, and a tweet's structural position within that propagation network carries information about its construct label that is orthogonal to the textual features alone. To capture this relational dimension and test whether the six constructs are structurally distinct, we applied four Graph Neural Network (GNN) architectures to a tweet-level propagation graph built from a dataset of tweets, with one tweet per construct.

\subsection{Graph Formulation}

The propagation graph is defined as:
\begin{equation}
G = (V, E, X),
\end{equation}
where each node $v_i \in V$ represents a single tweet, and each undirected edge $(v_i, v_j) \in E$ encodes a semantic propagation link. Node features $X$ were formed by concatenating a 3,000-dimensional TF-IDF bigram vector, using sublinear TF weighting and a minimum document frequency of 3, with a 6-dimensional construct-keyword hit-rate vector. The hit rate for each construct is defined as the fraction of its 17 defining keywords present in the tweet, providing a compact signal of construct affinity. All features were L2-normalized to unit length, yielding a total node-feature dimensionality of 3,006.

\subsection{Propagation Graph Construction}

Edges were generated through two complementary mechanisms. The first mechanism is semantic similarity. For every tweet, the $k = 15$ nearest neighbors by cosine distance in TF-IDF space were identified using a brute-force \texttt{NearestNeighbors} search. An undirected edge was retained for each pair whose cosine similarity exceeded the threshold $\tau = 0.15$. This threshold was selected to balance graph density with cross-construct noise; it produced a base graph with 474,992 undirected edges and an average node degree of approximately 26.4.

The second mechanism is keyword-driven. For each construct, all tweets containing at least two of that construct's defining keywords were identified. A local cosine $k$-nearest-neighbor subgraph, with $k = 5$, was then built within each such group and merged into the main graph. This dual-signal strategy ensures that the graph encodes both broad semantic similarity and targeted keyword-level construct affinity.

\subsection{GNN Architectures}

All four architectures were implemented in PyTorch Geometric v2.6.1. Each model was configured with three message-passing layers, batch normalization after every layer, and dropout with $p = 0.4$. Table~\ref{tab:gnn_architectures} summarizes the architectural configurations.

\begin{table}[t]
\caption{Graph Neural Network Architectures Used in the Propagation Analysis}
\label{tab:gnn_architectures}
\centering
\resizebox{\columnwidth}{!}{%
\begin{tabular}{p{0.18\columnwidth} p{0.55\columnwidth} r}
\toprule
\textbf{Model} & \textbf{Configuration} & \textbf{Parameters} \\
\midrule
GCN & Symmetric spectral normalization with a residual skip connection from the input layer to the first hidden layer. Hidden dimension: 256. & 1,608,199 \\
GAT & Two-layer attention mechanism. The first two layers used eight parallel attention heads with concatenated outputs to maintain a hidden size of 256; the output layer used a single head. & 839,445 \\
GraphSAGE & Mean-based inductive neighborhood aggregation, updating each node representation by concatenating its embedding with the mean of its neighbors' embeddings. & 1,675,271 \\
GIN & Weisfeiler--Leman expressive aggregation using a two-layer MLP, 256 to 256 with ReLU, applied to the sum of each node's own representation and its neighbors' representations. & 1,035,783 \\
\bottomrule
\end{tabular}%
}
\end{table}

\subsection{Training Protocol}

Each model was trained for up to 200 epochs with early stopping based on validation accuracy, using a patience value of 25 epochs. We used the AdamW optimizer with a learning rate of $5 \times 10^{-4}$ and weight decay of $5 \times 10^{-4}$. A cosine annealing schedule decayed the learning rate to $10^{-5}$ over the course of training. Label smoothing with $\epsilon = 0.1$ was added to the cross-entropy loss to reduce overconfidence, and gradient norms were clipped at 1.0.

The dataset was split into 80\% training, 10\% validation, and 10\% test sets using stratified sampling with a fixed random seed of 42. All experiments were run on CPU under PyTorch 2.8.0.

\subsection{Propagation Analysis}

Beyond classification, we conducted four structural analyses on the propagation graph:

\begin{enumerate}
    \item \textbf{Cross-construct edge density:} measuring the fraction of each construct's edges that connect to other constructs.
    \item \textbf{Degree centrality per construct:} assessing whether some constructs occupy more central positions in the propagation graph.
    \item \textbf{Greedy modularity community detection:} evaluating the community structure of the graph and the construct purity of each detected community.
    \item \textbf{t-SNE projection:} visualizing the GNN input feature space to assess construct separability.
\end{enumerate}

\subsection{Evaluation Metrics}

The four GNN models were assessed using the same metric suite applied to the preceding models: accuracy, Cohen's kappa, Matthews correlation coefficient (MCC), macro- and weighted-averaged F1, macro-averaged precision and recall, log loss, area under the receiver operating characteristic curve (AUC, one-vs-rest), and average precision. Per-class F1 scores were computed for each of the six social noise constructs, and a confidence analysis was performed by binning test predictions according to their maximum softmax probability.

\section{Results and Discussion}

\subsection{6X6 Model Performance}

The results of the 6 machine learning and deep learning models across the 6 social noise constructs are shown in the tables below. The 6 classification models are evaluated using both the social noise definitions and the definitions supplemented with keywords identified in previous studies as descriptive of social noise. The 6x6 comparative analysis produced 12 tables, 6 tables using social noise construct definitions only, and 6 tables using social noise construct definitions supplemented by keywords. Tables 2 and 4 show the results for RoBERTa classification with and without keywords. Tables 3 and 5 show the results for XGBoost classification model with keywords and without keywords. The results using the social noise definition have shown improvements and a good level of detection. The results using the social noise definition supplemented by keywords show much better results. This indicates that the system can benefit greatly from external knowledge structures that can be improved over time. Tables 6 and 7 show the performance of the 6 models across all social noise constructs using the social noise definition as well as the social noise constructs definition supplemented by additional keywords. 

\begin{table}[H]
\caption{RoBERTa classification results for each social noise construct. Overall accuracy = 97.17\%.}
\label{tab:roberta_transposed_macro}
\centering
\small
\begin{tabular}{lccccccc}
\toprule
\textbf{Metric} & 
\textbf{AP} & 
\textbf{CE} & 
\textbf{CA} & 
\textbf{IC} & 
\textbf{NB} & 
\textbf{RM} & 
\textbf{Macro Avg.} \\
\midrule
Precision        & 0.9715 & 0.9815 & 0.9591 & 0.9591 & 0.9797 & 0.9798 & 0.9718 \\
Recall           & 0.9650 & 0.9733 & 0.9767 & 0.9767 & 0.9667 & 0.9717 & 0.9717 \\
F1-score         & 0.9682 & 0.9774 & 0.9678 & 0.9678 & 0.9732 & 0.9757 & 0.9717 \\
Cohen's $\kappa$ & 0.9672 & 0.9698 & 0.9598 & 0.9589 & 0.9668 & 0.9710 & 0.9667 \\
MCC              & 0.9665 & 0.9670 & 0.9578 & 0.9567 & 0.9670 & 0.9716 & 0.9666 \\
\bottomrule
\end{tabular}
\end{table}
\FloatBarrier

\begin{table}[H]
\caption{XGBoost classification results for each social-noise constructs.}
\label{tab:xgboost_transposed_macro}
\centering
\small
\begin{tabular}{lccccccc}
\toprule
\textbf{Metric} & 
\textbf{AP} & 
\textbf{CE} & 
\textbf{CA} & 
\textbf{IC} & 
\textbf{NB} & 
\textbf{RM} & 
\textbf{Macro Avg.} \\
\midrule
Precision        & 0.9396 & 0.9796 & 0.9614 & 0.9672 & 0.9382 & 0.9523 & 0.9564 \\
Recall           & 0.9600 & 0.9600 & 0.9550 & 0.9350 & 0.9617 & 0.9650 & 0.9561 \\
F1-score         & 0.9497 & 0.9697 & 0.9582 & 0.9508 & 0.9498 & 0.9586 & 0.9561 \\
Cohen's $\kappa$ & 0.9487 & 0.9560 & 0.9498 & 0.9298 & 0.9289 & 0.9686 & 0.9476 \\
MCC              & 0.9454 & 0.9458 & 0.9431 & 0.9289 & 0.9279 & 0.9677 & 0.9476 \\
\bottomrule
\end{tabular}
\end{table}
\FloatBarrier

\begin{table}[H]
\caption{RoBERTa classification results without keywords for each social-noise constructs.}
\label{tab:roberta_no_keywords_transposed_macro}
\centering
\small
\begin{tabular}{lccccccc}
\toprule
\textbf{Metric} & 
\textbf{AP} & 
\textbf{CE} & 
\textbf{CA} & 
\textbf{IC} & 
\textbf{NB} & 
\textbf{RM} & 
\textbf{Macro Avg.} \\
\midrule
Precision        & 0.9210 & 0.9085 & 0.8620 & 0.8947 & 0.8195 & 0.8967 & 0.8837 \\
Recall           & 0.8950 & 0.9183 & 0.8850 & 0.8350 & 0.8700 & 0.9100 & 0.8856 \\
F1-score         & 0.9078 & 0.9134 & 0.8733 & 0.8638 & 0.8440 & 0.9033 & 0.8843 \\
Cohen's $\kappa$ & 0.8810 & 0.9001 & 0.8599 & 0.8089 & 0.8190 & 0.8871 & 0.8649 \\
MCC              & 0.8800 & 0.9101 & 0.8589 & 0.8000 & 0.8180 & 0.8710 & 0.8649 \\
\bottomrule
\end{tabular}

\par\smallskip
\begin{minipage}{0.95\linewidth}
\small
\textit{Note.} AP = Affiliation and Politics; CE = Conflict Engagement; CA = Cultural Agency; IC = Image Curation; NB = Norms and Beliefs; RM = Relationship Management. Overall accuracy = 88.69\%. RoBERTa was retrained from scratch on keyword-stripped text for three epochs using AdamW and $\eta = 2 \times 10^{-5}$. The model retained 91.3\% of its original accuracy.
\end{minipage}
\end{table}
\FloatBarrier

\begin{table}[H]
\caption{XGBoost classification results without keywords for each social-noise constructs.}
\label{tab:xgboost_no_keywords_transposed_macro}
\centering
\small
\begin{tabular}{lccccccc}
\toprule
\textbf{Metric} & 
\textbf{AP} & 
\textbf{CE} & 
\textbf{CA} & 
\textbf{IC} & 
\textbf{NB} & 
\textbf{RM} & 
\textbf{Macro Avg.} \\
\midrule
Precision        & 0.9435 & 0.8948 & 0.9043 & 0.9457 & 0.6805 & 0.8677 & 0.8727 \\
Recall           & 0.9183 & 0.8367 & 0.8817 & 0.9000 & 0.8733 & 0.7650 & 0.8625 \\
F1-score         & 0.9307 & 0.8648 & 0.8928 & 0.9223 & 0.7650 & 0.8131 & 0.8648 \\
Cohen's $\kappa$ & 0.9011 & 0.8267 & 0.8786 & 0.8981 & 0.6948 & 0.7278 & 0.8350 \\
MCC              & 0.9001 & 0.8261 & 0.8643 & 0.8991 & 0.6989 & 0.7274 & 0.8350 \\
\bottomrule
\end{tabular}

\par\smallskip
\begin{minipage}{0.95\linewidth}
\small
\textit{Note.} Overall accuracy = 86.25\%. XGBoost was retrained on keyword-stripped TF-IDF features with 1{,}000 trees, depth = 8, and $\eta = 0.05$. The model retained 90.2\% of its original accuracy. The drop in Norms and Beliefs precision reflects the construct's high keyword dependency.
\end{minipage}
\end{table}
\FloatBarrier

Table~\ref{tab:model-comparison} summarizes the comparative performance of the 6 classification models. RoBERTa attains the highest accuracy (97.17\%), accompanied by Cohen's $\kappa = 0.966$, MCC = 0.966, and macro F1 = 0.9717. These results underscore the effectiveness of pre-trained transformer-based representations for social-noise classification tasks. The BiLSTM model augmented with multi-head self-attention performs comparably, achieving 96.94\% accuracy ($\kappa = 0.963$, MCC = 0.963). This finding suggests that attention-enhanced recurrent architectures can approach transformer-level performance while operating with substantially fewer parameters. Meanwhile, the multi-scale residual CNN with squeeze-and-excitation attention reaches 96.69\% accuracy ($\kappa = 0.960$, MCC = 0.960), indicating that convolutional architectures remain competitive when strengthened by channel-wise attention mechanisms.

Among traditional machine learning methods, XGBoost achieves 95.61\% accuracy, indicating that gradient boosting applied to TF-IDF trigram features constitutes a strong baseline. Its performance reduces the gap with deep learning models to roughly 1.5 percentage points. In comparison, Random Forest achieves 91.69\% accuracy, approximately four percentage points lower than XGBoost, a difference attributable to the absence of sequential error correction inherent to bagging-based ensembles. AdaBoost performs considerably worse, with an accuracy of 79.94\%, highlighting the limitations of shallow weak learners and the sensitivity of exponential loss functions when modeling the complex feature space of social media text.

Further insight emerges from the log-loss values. RoBERTa exhibits the best probability calibration (0.1304), whereas AdaBoost records a substantially higher log loss of 0.600. This disparity suggests that AdaBoost tends to produce overconfident yet incorrect predictions, indicating weaker probabilistic reliability than the other models.

\begin{table}[H]
\caption{Text classification model performance comparison with keywords across the six social noise constructs.}
\label{tab:model-comparison}
\centering
\begin{tabular}{lcccccc}
\toprule
Model & Accuracy (\%) & Cohen's $\kappa$ & MCC & F1-score& Log Loss \\
\midrule
RoBERTa-base & 97.17 & 0.966 & 0.966 & 0.9717 & 0.1304 \\
BiLSTM+Attn & 96.94 & 0.963 & 0.963 & 0.9694 & 0.1270 \\
CNN+SE & 96.69 & 0.960 & 0.960 & 0.9670 & 0.1280 \\
XGBoost & 95.61 & 0.947 & 0.947 & 0.9561 & 0.1370 \\
Random Forest & 91.69 & 0.900 & 0.901 & 0.9169 & 0.3070 \\
AdaBoost & 79.94 & 0.759 & 0.761 & 0.8008 & 0.6000 \\
\bottomrule
\end{tabular}
\end{table}

To evaluate whether the classifiers capture the underlying semantic structure rather than relying on keyword labels, a keyword-removal ablation study was conducted. All keywords were removed using word-boundary regular expressions, such as \verb|\bgovernment\b|, and each model was retrained under identical conditions, including balanced sampling, stratified 80/20 splits, and a fixed random seed ($s=42$). The results in Table~\ref{tab:no-keywords} reveal a clear three-tier hierarchy of robustness.

In the top tier, Random Forest, RoBERTa, XGBoost, and CNN+SE retain approximately 90\% of their original accuracy, indicating strong resilience to the removal of explicit keyword signals. This pattern suggests that these models effectively leverage contextual and co-occurrence information beyond surface-level lexical cues. AdaBoost exhibits moderate performance degradation, reflecting its reliance on comparatively weaker feature representations. In contrast, BiLSTM with attention collapses to near-random performance (16.67\%, $\kappa = 0.000$), indicating strong dependence on keyword presence and limited generalization capacity in the absence of pre-trained contextual knowledge.

\begin{table}[H]
\caption{AI classification model performance comparison without keywords across the 6 social-noise constructs.}
\label{tab:no-keywords}
\centering
\begin{tabular}{lccccccc}
\toprule
Model & Acc. w/o KW & Acc. Drop & F1 w/o KW & F1 Drop & $\kappa$ w/o KW & $\kappa$ Drop & Retained \\
\midrule
RoBERTa & 88.69\% & $-8.48$ & 0.8870 & $-0.0847$ & 0.864 & $-0.102$ & 91.3\% \\
BiLSTM+Attn & 16.67\% & $-80.27$ & 0.0476 & $-0.9218$ & 0.000 & $-0.963$ & 17.2\% \\
CNN+SE & 87.06\% & $-9.63$ & 0.8735 & $-0.0935$ & 0.845 & $-0.115$ & 90.0\% \\
XGBoost & 86.25\% & $-9.36$ & 0.8648 & $-0.0913$ & 0.835 & $-0.112$ & 90.2\% \\
Random Forest & 83.83\% & $-7.86$ & 0.8415 & $-0.0754$ & 0.806 & $-0.094$ & 91.4\% \\
AdaBoost & 68.64\% & $-11.30$ & 0.7061 & $-0.0947$ & 0.624 & $-0.135$ & 85.9\% \\
\bottomrule
\end{tabular}
\end{table}

\begin{figure}[H]
\centering
\includegraphics[width=\textwidth]{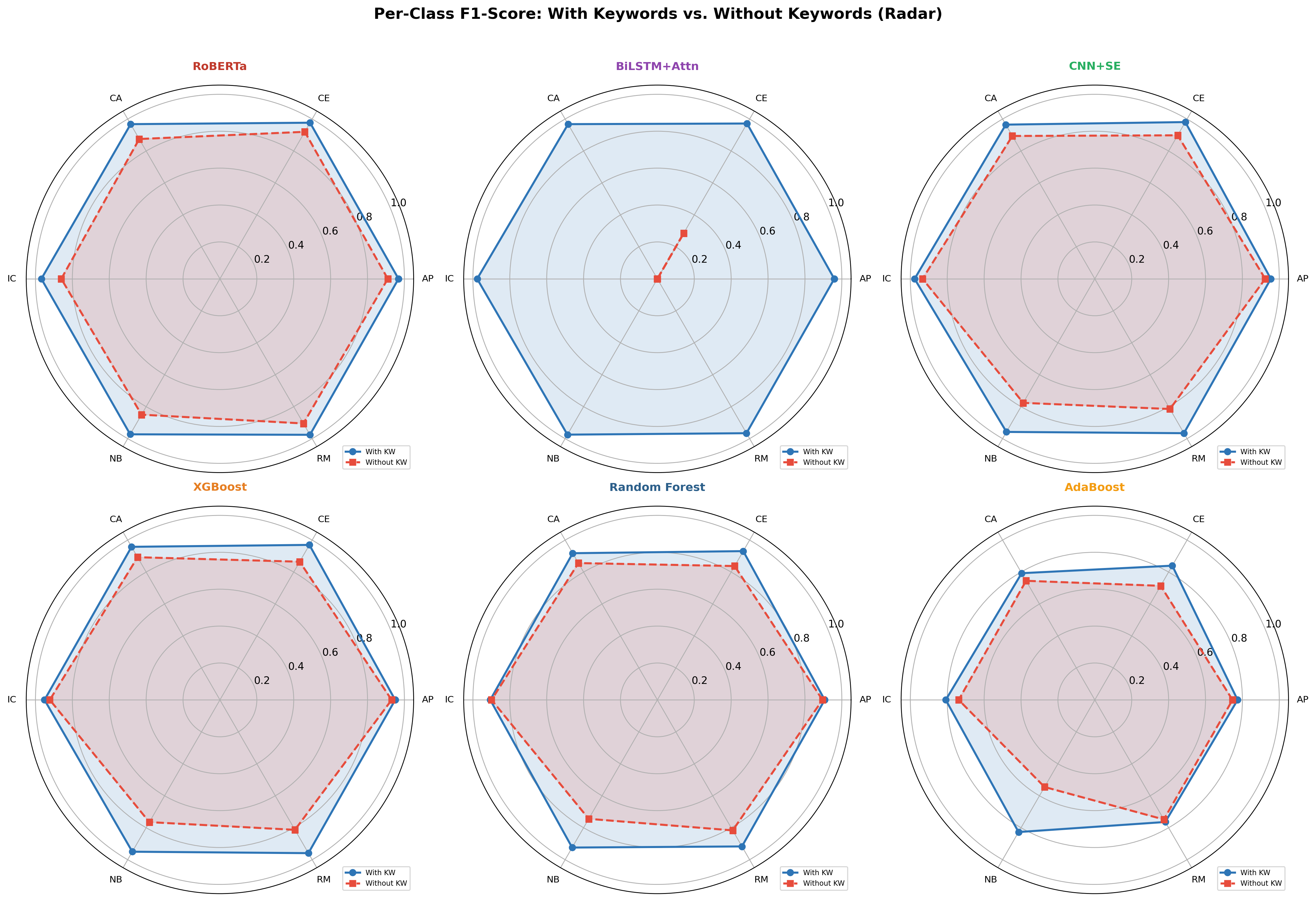}
\caption{social noise construct classification models comparison with keywords versus without keywords.}
\Description{Performance comparison of multiple classification models trained with and without keyword features, showing differences in accuracy and robustness.}
\label{fig:Performance Comparison After Keyword and before keyword for each six social noise constructs}
\end{figure}

Figure~\ref{fig:Performance Comparison After Keyword and before keyword for each six social noise constructs} presents radar plots comparing per-class F1-scores with keywords versus without keywords across all six models and six social-noise constructs. RoBERTa, CNN+SE, XGBoost, and Random Forest maintain near-congruent profiles after keyword removal, confirming robust contextual learning across constructs. BiLSTM+Attention collapses toward the origin, demonstrating strong keyword dependence in the absence of pre-trained representations. Norms and Beliefs exhibit the most pronounced inward contraction across the surviving models, consistent with the construct's keyword dependency identified in the LIME explainability analysis.

\subsection{Model performance convergence}

 The narrow performance range among the top text-based models is better explained by the task structure than by differences in model capacity. The six social noise constructs are highly separable at the lexical level, as reflected in near-perfect graph modularity ($Q = 0.9814$), which indicates strong alignment between semantic similarity and class membership. This degree of separability allows a wide range of models to arrive at similar decision boundaries. The balanced dataset further contributes to this consistency by removing class imbalance effects and ensuring equal representation during training.

At the same time, TF-IDF trigram features capture domain-specific vocabulary and recurring phrase patterns with sufficient precision to distinguish between constructs. Together, these factors create a setting in which performance quickly approaches a ceiling, and additional model complexity yields only marginal gains. This is evident in the relatively small improvement RoBERTa achieves over XGBoost, despite its substantially larger parameter space, illustrating diminishing returns in an already well-structured feature space. XAI analysis is therefore implemented to further justify the model.

\subsection{Explainable AI Analysis Using LIME}

The LIME analysis provides three forms of evidence for model trustworthiness. First, the models are not trivially memorizing keywords, as 89.6\% of prediction weight is derived from non-keyword features. Second, the learned features are semantically appropriate for each construct, such as political discourse features for Affiliation and Politics and humanitarian vocabulary for Relationship Management. Third, feature importance remains consistent across architecturally distinct models, suggesting that the observed patterns are properties of the data rather than artifacts of any specific learning algorithm.

Table~\ref{tab:xai-keyword} presents the explainability analysis conducted on the original dataset, where all 102 labeling keywords remain present in the text. For each correctly classified tweet, LIME for text models or Gradient $\times$ Input attribution for graph neural network models identified the top 20 most influential features, which were then categorized as either labeling keywords or non-keyword contextual words. The reported percentages indicate how much of each model's prediction weight is attributable to the labeling keywords.

Across all ten models, average keyword reliance ranges from 10.2\% for AdaBoost to 22.4\% for CNN+SE, demonstrating that even when keywords are available, models predominantly base their decisions on contextual features. At the construct level, Norms and Beliefs consistently exhibits the highest keyword reliance across paradigms, reaching 60.4\% for RoBERTa and 53.7\% for Random Forest. In contrast, Affiliation and Politics shows near-zero reliance on keywords across most models, confirming that political discourse is expressed through a broad vocabulary of geopolitical and institutional terms that extends beyond labeling keywords. CNN+SE presents an anomalous 96.7\% keyword weight for Conflict Engagement, attributable to its multi-scale convolutional filters strongly activating on conflict-specific $n$-grams such as ``war,'' ``attack,'' and ``military,'' which overlap with the labeling vocabulary.

\begin{table}[t]
\caption{XAI with keywords: keyword feature reliance (\%) across all ten models. Values indicate the percentage of each model's prediction driven by labeling keywords.}
\label{tab:xai-keyword}
\centering
\begin{tabular}{lrrrrrrr}
\toprule
Model & AP & CE & CA & IC & NB & RM & Avg. \\
\midrule
\multicolumn{8}{l}{\textit{Deep Learning Models}} \\
RoBERTa & 0.0 & 1.7 & 21.6 & 16.4 & 60.4 & 1.6 & 16.9 \\
BiLSTM+Attn & 6.7 & 20.0 & 1.2 & 2.1 & 45.6 & 19.3 & 15.8 \\
CNN+SE & 3.2 & 96.7 & 3.0 & 0.0 & 22.7 & 8.8 & 22.4 \\
\midrule
\multicolumn{8}{l}{\textit{Machine Learning Models}} \\
XGBoost & 0.9 & 18.0 & 0.0 & 0.0 & 34.4 & 9.1 & 10.4 \\
Random Forest & 0.0 & 27.7 & 0.0 & 0.0 & 53.7 & 15.1 & 16.1 \\
AdaBoost & 3.6 & 8.0 & 0.0 & 9.7 & 29.7 & 10.2 & 10.2 \\

\bottomrule
\end{tabular}
\par\smallskip
\begin{minipage}{0.95\linewidth}
\small
\textit{Note.} Keywords are present in the data. LIME was used for text models; Gradient $\times$ Input attribution was used for graph neural network models. Higher values indicate stronger reliance on keywords. $^{\dagger}$ Average computed over available constructs only. Dashes indicate insufficient correct predictions.
\end{minipage}
\end{table}

Table~\ref{tab:ablation} reports the results of a keyword-removal ablation experiment in which all labeling keywords were systematically removed using word-boundary regular expressions, and the six text classification models were retrained on the modified corpus. This setup isolates whether the models can still identify social-noise constructs when the defining vocabulary is absent.Graph neural network models are not included because removing keywords would require reconstructing the similarity graph, thereby altering its structure and making direct comparisons unreliable.

\begin{table}[t]
\caption{Performance without keywords: classification results after removing all labeling keywords.}
\label{tab:ablation}
\centering
\begin{tabular}{lcccccc}
\toprule
Model & Acc. w/o KW & Acc. Drop (pp) & F1 w/o KW & F1 Drop & $\kappa$ w/o KW & Retained \\
\midrule
\multicolumn{7}{l}{\textit{Deep Learning Models}} \\
RoBERTa & 88.69\% & $-8.48$ & 0.8870 & $-0.0847$ & 0.864 & 91.3\% \\
BiLSTM+Attn & 16.67\% & $-80.27$ & 0.0476 & $-0.9218$ & 0.000 & 17.2\% \\
CNN+SE & 87.06\% & $-9.63$ & 0.8735 & $-0.0935$ & 0.845 & 90.0\% \\
\midrule
\multicolumn{7}{l}{\textit{Traditional Machine Learning Models}} \\
XGBoost & 86.25\% & $-9.36$ & 0.8648 & $-0.0913$ & 0.835 & 90.2\% \\
Random Forest & 83.83\% & $-7.86$ & 0.8415 & $-0.0754$ & 0.806 & 91.4\% \\
AdaBoost & 68.64\% & $-11.30$ & 0.7061 & $-0.0947$ & 0.624 & 85.9\% \\
\bottomrule
\end{tabular}
\par\smallskip
\begin{minipage}{0.95\linewidth}
\small
\textit{Note.} All labeling keywords were removed via word-boundary regular expressions. Each model was retrained from scratch on keyword-stripped text using identical hyperparameters, balanced sampling (tweets per class), and a stratified 80/20 split (seed = 42). Acc. Drop is reported in percentage points (pp). Retained = accuracy without keywords $\div$ accuracy with keywords $\times 100$. Graph neural network models are excluded because keyword removal requires rebuilding the graph structure. BiLSTM+Attention collapsed to random chance for six equiprobable classes; $\kappa = 0.000$ confirms that predictions are statistically indistinguishable from random assignment.
\end{minipage}
\end{table}

The results reveal a clear three-tier pattern. Random Forest, RoBERTa, XGBoost, and CNN+SE retain roughly 90\% of their original accuracy, indicating that these models rely on broader contextual signals rather than keyword presence alone. This robustness is supported by either pre-trained representations, as in RoBERTa, or expressive feature spaces such as TF-IDF trigrams. AdaBoost shows a moderate drop, reflecting its more limited capacity. In contrast, BiLSTM+Attention collapses to near-random performance, indicating strong dependence on keywords as structural anchors.

Taken together with Table~\ref{tab:xai-keyword}, these results offer a complementary perspective. The explainability analysis shows how much the models rely on keywords when they are available, whereas the ablation study demonstrates whether meaningful classification remains possible without them. The combined evidence indicates that the six social-noise constructs are grounded in semantic patterns and that models with sufficient representational capacity can recover these patterns even in the absence of explicit keyword cues.

\begin{figure}[t]
\centering
\includegraphics[width=\textwidth]{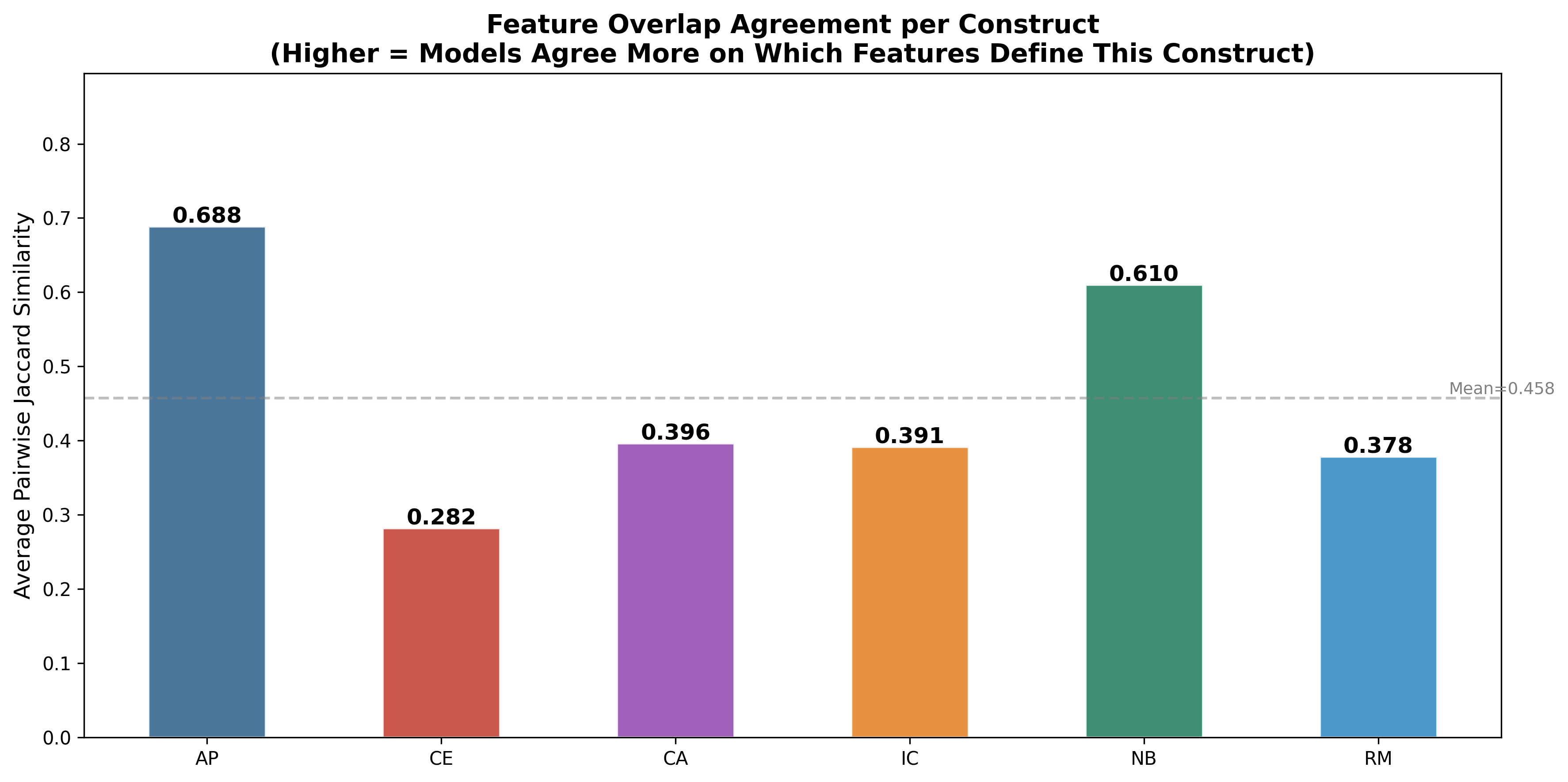}
\Description{Bar chart showing average pairwise Jaccard similarity of top-15 LIME features across four models for each of the six social noise constructs. Affiliation and Politics have the highest overlap at 0.688, followed by Norms and Beliefs at 0.610, while Conflict Engagement has the lowest at 0.282.}
\caption{Average pairwise Jaccard similarity of top-15 LIME features across four models (RoBERTa, XGBoost, Random Forest, AdaBoost) for each social noise construct. Higher values indicate greater cross-model agreement on which features define a given construct. The dashed line marks the overall mean ($J = 0.458$).}
\label{fig:jaccard}
\end{figure}

Figure~\ref{fig:jaccard} quantifies the degree to which architecturally distinct models converge on the same discriminative features for each social-noise construct. Affiliation and Politics exhibits the strongest cross-model agreement ($J = 0.688$), indicating that political discourse in the \#Ukraine corpus is characterized by stable contextual markers, including geopolitical references, institutional terminology, and policy-related framing. Norms and Beliefs follow closely ($J = 0.610$), consistent with its specialized religious and moral vocabulary. The remaining constructs cluster between $J = 0.282$ and $J = 0.396$, with Conflict Engagement showing the lowest agreement. The overall mean Jaccard similarity of 0.458 falls above what would be expected if models selected features at random from the shared vocabulary, providing quantitative evidence that the six constructs possess model-independent semantic signatures rather than architecture-specific artifacts.

\FloatBarrier

\subsection{Practical Implications}

The findings of this study carry several implications for practitioners in content moderation, platform design, and computational social science.

First, the improved classification accuracy achieved by multiple models, with four models exceeding 95\%, demonstrates that automated social-noise detection is technically feasible and can be deployed at scale to filter or label noisy content in real-time social media streams. The availability of strong performance from both deep learning models and traditional machine learning models provides flexibility in deployment. Organizations with limited computational resources can deploy XGBoost with TF-IDF features, incurring only a 1.56 percentage-point reduction in accuracy relative to the transformer model.

Second, the 6 social noise knowledge structures provide a more nuanced framework for content moderation than binary noise-versus-signal classification. Rather than simply removing noisy content, platform designers can use construct-level classifications to implement differentiated responses. Casual expressions may be deprioritized in information-seeking feeds but retained in social interaction contexts; rumor-related content may trigger fact-checking prompts; and appeals to popularity may be flagged with contextual warnings about bandwagon reasoning. This construct-aware approach preserves the diversity of user communication styles while targeting the specific types of noise that degrade information quality.

Third, the explainability results provide a foundation for building trustworthy and auditable classification systems. The finding that high-performing models derive most of their decision weight from non-keyword features addresses concerns about superficial pattern matching and indicates that the models learn meaningful semantic representations. Combined with a confidence-based human-in-the-loop threshold, organizations can deploy these systems with appropriate oversight by routing uncertain predictions to human reviewers while processing the majority of content automatically.

\subsection{GNN Model Evaluation}

Table~\ref{tab:gnn_full_evaluation} reports the full evaluation of all four GNN models on the held-out test tweets.

\begin{table}[t]
\caption{Full Evaluation of the Four GNN Models on the Held-Out Test Set}
\label{tab:gnn_full_evaluation}
\centering
\resizebox{\columnwidth}{!}{%
\begin{tabular}{lrrrr}
\toprule
\textbf{Metric} & \textbf{GCN} & \textbf{GAT} & \textbf{GraphSAGE} & \textbf{GIN} \\
\midrule
Accuracy & 0.3010 & 0.6362 & 0.3514 & 0.6452 \\
Cohen's $\kappa$ & 0.1844 & 0.5756 & 0.2433 & 0.5861 \\
MCC & 0.2219 & 0.5759 & 0.2819 & 0.5867 \\
F1 Macro & 0.1942 & 0.6377 & 0.2756 & 0.6471 \\
F1 Weighted & 0.1942 & 0.6377 & 0.2756 & 0.6471 \\
Precision Macro & 0.2756 & 0.6408 & 0.4188 & 0.6528 \\
Recall Macro & 0.3010 & 0.6362 & 0.3514 & 0.6452 \\
Log Loss & 1.9268 & 1.0958 & 1.9245 & 1.0737 \\
AUC OvR & 0.7504 & 0.8942 & 0.8779 & 0.8989 \\
Average Precision & 0.3964 & 0.7097 & 0.6593 & 0.7238 \\
\bottomrule
\end{tabular}%
}
\end{table}

The Graph Isomorphism Network (GIN) led on every primary metric, achieving 64.52\% accuracy, $\kappa = 0.5861$, MCC = 0.5867, macro-F1 = 0.6471, and AUC = 0.8989. According to the Landis--Koch interpretation scale, this $\kappa$ value indicates substantial agreement~\cite{landis1977measurement}. The Graph Attention Network (GAT) followed closely, with 63.62\% accuracy and AUC = 0.8942. By contrast, GCN and GraphSAGE achieved substantially lower accuracies of 30.10\% and 35.14\%, respectively. This sharp split, with attention-based and MLP-based architectures outperforming the spectral and mean-aggregation models by roughly 30 percentage points, suggests that the structure of the social noise propagation graph is not well captured by uniform neighborhood averaging.

Two compounding effects help explain this gap. The first is over-smoothing. At an average node degree of 26.4, three rounds of uniform mean aggregation in GCN's spectral convolution gradually pulled node representations toward a class-indistinct global average. The residual skip connection in our GCN implementation dampened this effect but did not fully overcome it, as reflected by the model's early stopping at epoch 36 and its final accuracy of 30.10\%, only marginally above the 14.3\% random-chance baseline for a seven-class classification task.

The second effect is cross-construct social noise amplification. Because 23--31\% of edges span construct boundaries, GraphSAGE's mean aggregation absorbed misleading signals from neighboring tweets belonging to other constructs. GAT behaved differently: its learnable multi-head attention coefficients, implemented with eight heads, assigned near-zero weight to cross-construct edges, effectively performing soft neighborhood pruning. GIN's MLP-based aggregation likewise preserved the fine-grained structural fingerprint of each node's local subgraph, allowing it to distinguish construct-specific propagation patterns that averaging-based approaches tend to erase.

\subsection{Per-Class F1 Analysis}

Table~\ref{tab:gnn_per_class_f1} reports per-class F1 scores for all four models across the six social noise constructs and the Uncategorized class.

\begin{table}[t]
\caption{Per-Class F1 Scores Across the Seven Classification Categories}
\label{tab:gnn_per_class_f1}
\centering
\resizebox{\columnwidth}{!}{%
\begin{tabular}{lrrrr}
\toprule
\textbf{Construct} & \textbf{GCN} & \textbf{GAT} & \textbf{GraphSAGE} & \textbf{GIN} \\
\midrule
Affiliation and Politics & 0.0641 & 0.6319 & 0.2471 & 0.6431 \\
Conflict Engagement & 0.3958 & 0.6992 & 0.0000 & 0.6947 \\
Cultural Agency & 0.0000 & 0.6309 & 0.2286 & 0.6083 \\
Image Curation & 0.0000 & 0.6609 & 0.4114 & 0.6784 \\
Norms and Beliefs & 0.4972 & 0.6929 & 0.6441 & 0.6990 \\
Relationship Management & 0.0000 & 0.6271 & 0.0000 & 0.6618 \\
\bottomrule
\end{tabular}%
}
\end{table}

GIN and GAT were consistently strong across all six constructs, with F1 scores ranging from 0.6083 for Cultural Agency under GIN to 0.6992 for Conflict Engagement under GAT. Conflict Engagement achieved the highest F1 score in both leading models, with GAT reaching 0.6992 and GIN reaching 0.6947. This result reflects the distinctive lexical and propagation signature of conflict-related tweets, including military terminology, strong emotional valence, and tight semantic clustering in the graph. Norms and Beliefs ranked second under GIN, with an F1 score of 0.6990, suggesting that ideological and moral-reasoning tweets form internally cohesive propagation clusters even in a semantically derived graph.

The failures of GCN and GraphSAGE on particular constructs are also diagnostically important. Relationship Management scored F1 = 0.0000 under both models, meaning that every tweet in this construct was misclassified. This failure is likely driven by vocabulary overlap: solidarity-oriented terms such as ``support,'' ``help,'' ``together,'' and ``humanitarian'' appear across multiple constructs, producing a high cross-construct edge ratio that uniform aggregation cannot untangle. Cultural Agency and Image Curation likewise scored F1 = 0.0000 under GCN, confirming that spectral convolution collapses these constructs' representations into the dominant class signal.

The Uncategorized class scored lowest under both GAT and GIN, with F1 scores of 0.5207 and 0.5444, respectively. This result is expected because the category contains tweets without a clear construct-specific keyword signature, making them inherently harder to propagate through construct-specific graph channels.

\section{Conclusion}

This study evaluated the effectiveness of contemporary natural language processing, deep learning, and graph-based learning approaches for detecting social noise in social media discourse. Building upon prior theoretical work, social noise was operationalized through six constructs—Affiliation and Politics, Conflict Engagement, Cultural Agency, Image Curation, Norms and Beliefs, and Relationship Management—and implemented through a keyword-enhanced classification framework. Using a large corpus of tweets collected from the \#Ukraine hashtag, the study conducted a comprehensive 6×6 matrix evaluation of machine learning, deep learning, and four graph neural network models. The results demonstrate that theoretically derived social noise constructs can be successfully leveraged for automated detection. Among the text-based models, RoBERTa achieved the highest performance (97.17

Network analysis provided additional evidence supporting the validity of the proposed framework. The social noise constructs exhibited strong structural coherence, reflected by exceptionally high modularity (Q = 0.9814), suggesting that the identified categories represent distinct yet interconnected dimensions of social discourse. Explainability analyses further demonstrated that high-performing models rely primarily on contextual, semantic, and pragmatic features rather than simple keyword matching. An important finding emerged from the graph-based experiments. Although GIN achieved lower classification accuracy (64.52

The practical implications of this work are significant for content moderation, platform design, and computational social science. The availability of high-performing models with varying computational requirements enables flexible deployment across institutional settings. In addition, the granular taxonomy supports more nuanced moderation strategies beyond binary filtering, while the confidence-based human-in-the-loop framework balances automated efficiency with appropriate human oversight for uncertain predictions.

\section{Future Work}

Future work will focus on investigating the performance of large language models in using social noise constructs across multiple datasets and social media platforms as an effective method in mitigating the effect of misinformation. Comparative studies involving instruction-tuned and reasoning-enhanced AI algoruthms and models may provide valuable insights into their ability to capture subtle social, cultural, and contextual signals that extend beyond traditional supervised learning approaches. Future work should evaluate the generalizability of the proposed framework across diverse datasets, social media platforms, and communication contexts. The findings from this study suggest significant opportunities for deeper integration of textual and network-based representations. While transformer-based models demonstrated superior classification accuracy, graph neural networks captured complementary structural information embedded within social interactions and content propagation patterns. A particularly promising direction is the design of heterogeneous graph transformers capable of simultaneously encoding tweet text, user relationships, engagement patterns, and propagation structures within a unified framework. Such architectures could leverage attention mechanisms across multiple node and edge types, enabling richer representations of social noise phenomena.

\bibliographystyle{ACM-Reference-Format}
\bibliography{references2}
\newpage
\appendix

\section{Appendix}

\begin{table}[h]
\caption{Social Noise Constructs definitions and sample keywords}
\label{tab:social-noise-constructs}
\centering
\footnotesize
\renewcommand{\arraystretch}{1.15}
\setlength{\tabcolsep}{4pt}
\begin{tabularx}{\textwidth}{
>{\RaggedRight\arraybackslash}p{0.17\textwidth}
>{\RaggedRight\arraybackslash}p{0.25\textwidth}
>{\RaggedRight\arraybackslash}X}
\toprule
\textbf{Construct} & \textbf{Definition} & \textbf{All Keywords} \\
\midrule

Image Curation 
& Is the effort by a social media user, consciously or unconsciously, to craft their online identities 
& please, willing, identity, recommend, report, thank, determination, express, filter, glow up, online persona, blessed, appreciate, grateful, effort, try, connect, inspire, style, follow, vibe, share, careful, build, personalize, aesthetic, rebrand, curate, highlight, showcase, authentic, unfiltered, realme, selfie, transformation, glow, reinvent, portfolio, brand, influencer, goals, milestone, achievement, flex, posture, visibility, credibility, impression, perception, polished, reputation, narrative, self-promote, presence, lifestyle, snapshot, moment, debut, reveal, spotlight \\
\addlinespace

Relationship Management 
& Refers to a user's understanding of their roles and responsibilities within social institutions as well as the level of confidence in their personal beliefs 
& love, want, help, please, equivalent, humanity, stop, violence, awareness, defense, influence, mutual aid, absurdity, ordained, advising, trust, personal, values, support, system, role, follow, link, caretaker, share, build, bond, network, community, solidarity, ally, engage, connect, collaborate, partnership, unity, belong, inclusion, outreach, brotherhood, sisterhood, tribe, circle, uplift, empower, mentor, peer, kinship, loyalty, reciprocity, validate, affirm, acknowledge, compassion, empathy, check in, hold space, stand with, together, with you \\
\addlinespace

Conflict Engagement 
& Is the level of social conflict with which a user is comfortable 
& report, stop, humiliation, mistreat, misconduct, controversial, discord, confront, stand, call out, cancel, criticism, drama, argue, debate, disagree, feud, end, stand by, challenge, exposed, push back, clash, trigger, offensive, toxic, harass, troll, bully, attack, block, mute, ban, dragged, ratio, receipts, problematic, accountability, blast, cancel culture, beef, clap back, shut down, respond, dispute, resist, counter, reject, silence, defend, expose, unfollow, clapback, petty \\
\addlinespace

Cultural Agency 
& Is characterized by civic participation in social issues and is exhibited by individuals who believe in their own power to be heard and to shape culture and beliefs 
& radical, recommend, demand, justice, lives, matter, revolution, leadership, diversity, community, social, systemic, engagement, civic, raise, awareness, empowered, movement, voice, agency, disrupt, change, advocate, cultural, shift, representation, take, stand, allyship, protest, activism, mobilize, solidarity, march, petition, boycott, organize, grassroots, narrative, reclaim, decolonize, equity, liberation, amplify, platform, intersectionality, privilege, oppression, systemic change, reform, rally, coalition, impact, visibility, humanize, uplift, marginalized, speak up, movement building, social justice, power \\
\addlinespace

Affiliation and Politics 
& Is characterized by loyalty to a political party, religion, or an organization. This could include people paid to carry out certain activities or advertise certain products. 
& Foxnews, sisracist, systemic, racism, supremacist, supremacy, racist, white, left, fascist, abiding, loyalist, influencer, vote, supporter, endorse, propaganda, party line, lobby, affiliated, news, report, campaign, black, right, MAGA, woke, liberal, conservative, leftist, right-wing, Democrat, Republican, partisan, polarized, deep state, mainstream media, fake news, conspiracy, agenda, endorsed, sponsored, bot, astroturfing, dark money, PAC, echo chamber, tribe, dogma, manifesto, ideology, regime, coup, fringe, extremist, radicalized, red pill, blue pill, disinformation, influence operation \\
\addlinespace

Norms and Beliefs 
& Is characterized by deep beliefs, culture, ideology, religion, a cause etc. 
& change, culture, demand, justice, equity, inclusion, fight, racism, socialists, communists, anarchists, conservatives, extremists, religion, moral, values, cultural, identity, spiritual, practices, tradition, faith, doctrine, sacred, afterlife, scripture, nationalism, karma, virtue, ethics, principle, creed, conviction, righteousness, worldview, taboo, sin, blessed, prayer, amen, truth, awakening, salvation, prophet, divine, judgment, heresy, redemption, blasphemy, anti-woke, natural law, evolution, creationism, pro-life, pro-choice, freedom, sovereignty, anarchy, patriotism, censorship, heritage, purity, indoctrination \\
\bottomrule
\end{tabularx}
\end{table}
\FloatBarrier
\end{document}